\pdfoutput=1
\documentclass[journal]{IEEEtran}
\usepackage{cite}
\usepackage{amsmath,amssymb,amsfonts}
\usepackage{algorithmic}
\usepackage{graphicx}
\usepackage{textcomp}
\usepackage{booktabs}

\begin{document}

\title{BnBERT-iPET: Sparse Few-Shot Language Modeling for Bengali via Lottery Ticket Pruning}

\author{Sajib~Hossain,
        Md~Kamrus~Samad,
        Anan~Ghosh,
        Labib~Imam~Chowdhury,
        and~Nabeel~Mohammed%
\thanks{The authors are with the Department of Electrical and Computer Engineering, North South University, Dhaka, Bangladesh.}%
\thanks{Corresponding author: Sajib Hossain (e-mail: sajib.hossain03@northsouth.edu).}}

\markboth{Preprint}%
{Hossain \MakeLowercase{\textit{et al.}}: BnBERT-iPET: Sparse Few-Shot Language Modeling for Bengali}

\maketitle

\begin{abstract}
Deep neural networks have shown impressive success in NLP tasks owing to their complex structure and huge number of edges. Achieving state-of-the-art performance in natural language processing with a large pre-trained model such as BERT is expensive and time-consuming, carries a large carbon footprint, and is difficult to realize on machines with minimal computational capability. This creates a barrier to training complex models for resource-constrained languages such as Bengali. However, in a complex neural model, not all edges are equally impactful, and the contributions of some of them can be neglected. Pruning promises to reduce the memory footprint of regular networks, shorten the training time of ever-growing networks, and increase inference efficiency without sacrificing comparable performance. In this work, we introduce BnBERT-iPET, a sparse few-shot language modeling approach for Bengali, and experimentally show that a lightweight few-shot-learned language model retaining only 10\% of the edges of an initial model such as BERT can perform neck and neck with much larger models on challenging tasks for a resource-constrained language such as Bengali. By learning from few shots through iterative pattern exploiting training and achieving 90\% sparsity with the Lottery Ticket Hypothesis pruning technique, our pruned BnBERT-iPET model proves to be a tough competitor to state-of-the-art language models such as Bangla Electra, Indic-BERT, and XLM-RoBERTa on downstream tasks over standard benchmark datasets of the Bengali language.
\end{abstract}


\section{Introduction}
\label{sec:introduction}
\IEEEPARstart{T}{he} use of computational techniques in natural language processing enables the learning, understanding, and creation of material in human languages\cite{b1}. In recent years, natural language processing (NLP) has seen revolutionary developments. However, it is not as equitable as it would seem to develop and apply NLP technology. The dataset size is not an issue for high-resource language pairings (such as English and French) because researchers have acquired a large number of parallel corpora through reliable sources. But the necessity of having large amounts of parallel data is not a reasonable assumption for almost 7000+ languages that are now in use worldwide. Thus, it is seen as a significant challenge for low-resource languages\cite{b2}. However, remarkable achievements have been made in many applications as a result of the creation and implementation of pre-trained language models. Pre-trained language models have the significant benefit of being universal language processing tools. They already contain a large portion of the knowledge required for language processing, so only a small amount of labeled data is needed to fine-tune a pre-trained language model\cite{b3}. Despite having nearly 230 million native speakers and being one of the most widely spoken languages in the world, Bangla is still ranked as a low-resource language by the natural language processing (NLP) community\cite{b4}. Research on Bangla Natural Language Processing (BNLP) started in the early 1990s\cite{b5} and later continued for Parts of Speech (POS) tagging \cite{b6}, Named Entity Recognition (NER) \cite{b7}, sentiment analysis \cite{b8}, emotion detection \cite{b9}, and news categorization \cite{b10, b12}. There is a survey on Bangla Natural Language Processing (BNLP) that presented a thorough analysis of 75 BNLP research papers that appeared from 1999 to 2021 and categorized them into 11 categories based on Classical, Machine Learning, and Deep Learning methods\cite{b11}.

Natural language processing changed dramatically when the research community introduced a new language representation model called BERT (Bidirectional Encoder Representations from Transformers) \cite{b13}, which achieved state-of-the-art performance. It performs extremely well on the tasks mentioned earlier, but it requires enormous resources to run, which is a major problem and a visible challenge for low-resource computing machines. The BERT base model uses 12 layers of transformer blocks with a hidden size of 768 and 12 self-attention heads, totaling around 110M trainable parameters, while the large version uses 24 layers of transformer blocks with a hidden size of 1024 and 16 self-attention heads, totaling around 340M trainable parameters; both are costly to train and deploy. Facebook AI later robustly optimized the BERT pretraining approach \cite{b14}. For resource-limited languages like Bangla in particular, such models are complex and complicated to train and deploy, small models with fewer layers suffer from performance issues, and big networks demand more resources, leading to a considerable carbon footprint and environmental impact. To address this problem, iPET \cite{b22}, a parameter- and language-model-optimization method, can outperform big language models such as GPT-3, which needs 175 billion parameters, on various tasks by converting textual inputs into cloze questions that contain a task description, combined with gradient-based optimization; exploiting unlabeled data gives further improvements. In addition, DistilBERT \cite{b15} leverages knowledge distillation during the pre-training phase, making the model up to 40\% lighter. Many other methods exist for making models lighter, such as compressing neural networks with the hashing trick \cite{b16}, gradient- and magnitude-based pruning for sparse deep neural networks \cite{b17}, pruning filters for efficient ConvNets \cite{b18}, channel pruning for accelerating very deep neural networks \cite{b19}, and deep compression with pruning, trained quantization, and Huffman coding \cite{b20}. Among them, one of the most striking works on making models lighter and smaller is the Lottery Ticket Hypothesis \cite{b21}, which is based on finding sparse, trainable subnetworks and dropping unnecessary connections.\\ 
To address this issue, we developed a single, diversified dataset for the Bangla language, which includes content from Bangla newspapers, social media, and Shadhu bhasha. Furthermore, we introduce three distinct models, named BnBERT, BnBERT iPET, and BnBERT iPET Pruned. The initial model, BnBERT, has been fine-tuned on our diverse dataset and is capable of few-shot learning through Iterative Pattern Exploiting Training (iPET)\cite{b22}. This is achieved by converting textual inputs into cloze questions that contain a task description, combined with gradient-based optimization; exploiting unlabeled data gives further improvements. Additionally, we use the pruning technique from the Lottery Ticket Hypothesis\cite{b21} to remove almost 90\% of the negligible weights from the network while maintaining accuracy. To evaluate the potential of our models and compare the outcomes with those of other well-known models benchmarked on the same downstream tasks, the variety of downstream tasks in Bangla considered in this study was deliberately constrained.\\
Our contributions are summarized as follows:

\begin{itemize}
  \item We present a diversified Bangla unsupervised language dataset (BanglaDDS) for language modeling.
  \item We introduce three new models: BnBERT, BnBERT iPET, and BnBERT iPET Pruned.
  \item This research applies iterative pattern exploiting training, which makes BnBERT iPET capable of few-shot learning in the Bangla language.
  \item We propose the BnBERT iPET Pruned model, which is trained on smaller subnetworks instead of the entire model without sacrificing performance, using the Lottery Ticket Hypothesis pruning method to reach 90\% sparsity.  
\item We measured the perplexity of our three models and of each part of the dataset individually to evaluate how well a probability model can predict a given sample.
\item This experiment compares the performance and computational-complexity trade-offs between existing transformer-based models and the BnBERT iPET Pruned model.
\item We evaluated the three models on six downstream natural language processing tasks: sentiment classification, punctuation restoration, POS tagging, news categorization, authorship classification, and emotion classification. Moreover, performance comparisons with other transformer-based models are reported for these downstream tasks.

\end{itemize}

\section{Literature Review}
\vspace{2mm}
\label{sec:Literature review}
Although NLP research has advanced greatly in high-resource languages, it is still in its infancy for languages with limited resources, such as Bengali. Although Bangla is a low-resource language with a dearth of NLP-related research, considerable work on Bangla Natural Language Processing has continued for Parts of Speech (POS) tagging, Named Entity Recognition (NER), sentiment analysis, emotion detection, and news categorization. There is a comparison of the performance of a few statistical (n-gram, HMM) and transformation-based (Brill's tagger) POS tagging methods for the Bangla language \cite{b6}. In order to tag correctly, a supervised POS tagging system needs a sizable volume of the annotated training corpus. However, annotated corpus resources in Bangla are quite scarce. The POS tagging performance for Bangla is getting close to that of English with the use of N-gram (unigram), HMM, and Brill's transformation-based approaches. Performance was 55\% using a training set of about 5000 words and 41 tags. To categorize emotions in Bengali literature in the absence of a standard corpus, a corpus named ``BEmoC'' was created \cite{b10}. This corpus saw a significant rise in all scores after using transformer-based models. Bangla-BERT has the lowest F1 score 61.91\% of the transformer-based models. The XLM-R model significantly outperforms Bangla-BERT and m-BERT by roughly 6\% and 5\%, respectively. It obtained the highest F1 score of any model, coming in at 69.73\%. A performance evaluation of BEmoC revealed that, of all the techniques, the transformer model XLM-R produced the best results. In particular, XLM-R had the highest F1 score, coming in at 69.61\%. There is also a comparison for two pre-trained transformer models, multilingual BERT and XLM-RoBERTa (with fine-tuning), for sentiment analysis for the Bengali language \cite{b23}. Fine-tuned multilingual BERT and XLM-RoBERTa reached accuracies of 63\%--94\% and 68\%--95\%, respectively, yielding state-of-the-art results. Also, there is an augmentation strategy and exploration of different transformer-based models for the Punctuation Restoration task, focusing on high-resource (English) and low-resource (Bangla) languages \cite{b24}. This strategy used bidirectional Long Short-Term Memory on top of the pre-trained transformer network and obtained the best result using the XLM-RoBERTa model as it is trained with more texts for low-resource languages like Bangla and has a larger vocabulary for them. The performance gain from augmentation is marginal, but it helps human labelers reduce their time and effort and makes the manual annotation process faster. Finally, there is a publication for reviewing Bangla NLP tasks, resources, and tools accessible to the research community \cite{b12}; next, benchmarking datasets acquired from multiple platforms for nine NLP tasks using current state-of-the-art algorithms (i.e., transformer-based models). There is a comparison between monolingual vs. multilingual models of various sizes to obtain comparative outcomes for the NLP tasks under consideration. After performing 175 different experiments, findings suggest that transformer-based models work well, but that there is a trade-off in terms of computing expenses.\\ 
By providing a pretrained language model with natural-language ``task descriptions'', some NLP tasks can be performed entirely unsupervised. PET (Pattern Exploiting Training) is a semi-supervised training method that combines the ideas of providing task descriptions to pre-trained language models and standard supervised training by reformulating input instances as cloze-style phrases\cite{b25}. For a variety of tasks and languages, PET greatly surpasses supervised training and strong semi-supervised techniques in low-resource contexts. PET also promises few-shot learning for typically sized models, allowing input examples in the form of cloze-style phrases even when the number of examples is limited. By converting textual inputs into cloze questions that contain a task description, paired with gradient-based optimization and the exploitation of unlabeled data, the principle of PET is used to obtain parameter-efficient language models that can outperform massive language models like GPT-3, which needs 175 billion parameters, on various tasks\cite{b22}. This method is known as iPET (Iterative Pattern Exploiting Training). iPET, which utilizes ALBERT and multiple generations of models trained on datasets of increasing size that have been labeled by earlier generations, can outperform GPT-3 in a variety of tasks while using only 0.1\% of GPT-3's parameters. With LMs that have three orders of magnitude fewer parameters than GPT-3 on SuperGLUE, it is possible to achieve a few-shot text classification performance that is comparable to that of GPT-3.
Alongside achieving state-of-the-art performance across a wide variety of natural language tasks, the size of these models, their training cost, the required computational resources, and their latency have increased significantly as well. A very powerful and familiar method to make a model lighter is pruning. Most pruning techniques are unstructured or block-structured. Weight-based pruning removes weights individually, producing sparse matrices across the network \cite{b28} that are difficult to support on most hardware. As a solution to this problem, a generic, improved, and more efficient structured method based on adaptive low-rank factorization \cite{b29} preserves the fully dense structure of the weight matrices, eliminating the need for special linear algebra primitives and hardware for computational speedup. Compared to row- and column-based pruning, low-rank factorization better preserves the linear transformation of the uncompressed matrices. This suggests that pruning a large model is consistently better than training a small model from scratch, and that low-rank-based pruning achieves better performance than removing matrix columns and input features. The performance of FLOP-AGP, especially in comparison with its unstructured counterpart AGP, highlights the effectiveness of factorization-based pruning. The FLOP-$l_0$ method achieved an outstanding test perplexity (PPL) of 25.3---a loss of only 0.8 perplexity---while removing 50\% of the model parameters, as the base model adopted adaptive word embedding and softmax layers that already reduce the model size significantly. The work also reports the overall size of the recurrent encoder and adaptive embedding layers at the compression levels tested, and breaks down parameter usage within three word clusters based on frequency; FLOP learns to prune dimensions more aggressively for less-frequent words. Models like BERT, XLNet, and T5 are notable names in the field of NLP\cite{b30}. The larger the model, the more transformer blocks and self-attention heads are packed into the network, which increases model complexity and training time. Hessian-based ultra-low-precision quantization of BERT\cite{b31}, patient knowledge distillation for compressing a large BERT model into an equally effective lightweight shallow network \cite{b32}, and evaluating whether ``winning ticket'' initializations exist in the domains of natural language processing (NLP) and reinforcement learning (RL)\cite{b33} are all directed at network compression. The work of Frankle \& Carbin \cite{b21} thus surprised many researchers by presenting a simple algorithm for finding sparse subnetworks within larger networks that are trainable from scratch: set all weights smaller than some threshold to zero, prune them, and rewind the remaining weights to their initial configuration; then retrain the network from this starting configuration with the zeroed weights frozen (not trained). Another research work deconstructed the lottery ticket algorithm, focusing on three major and very critical components that can be varied significantly without impacting the overall results \cite{b34}, modifying the method while yielding impactful results for network pruning.

In recent research work, transfer learning based on hybrid deep learning models like LSTM, CNN, and CRF in NER is employed to assess Bangla-BERT, a monolingual BERT model for the Bangla language\cite{b26}. Using 40 GB of text data, BanglaLM, the biggest dataset for a Bangla language model, is employed for its pre-training. The initial stage of pre-training is the creation of a vocabulary utilizing the available corpora. Furthermore, byte-pair encoding (BPE) is mostly used to construct the cased and uncased vocabularies.
Apart from that, BanglaBERT and BanglishBERT are two NLU models for Bangla, a widely spoken yet low-resource language\cite{b27}. BanglaBERT is a BERT-based Natural Language Understanding (NLU) model that has been pretrained in Bangla on 27.5 GB of data from 110 well-known Bangla websites. BanglishBERT is a model pretrained in both languages that allows zero-shot transfer learning between Bangla and English. While most downstream task datasets for NLP applications are in English, this research work provides datasets for two hitherto underexplored tasks in Bangla: Natural Language Inference (NLI) and Question Answering (QA).

\section{Proposed Methodology}
\begin{figure}[!t]
\centering
\includegraphics[width=\columnwidth]{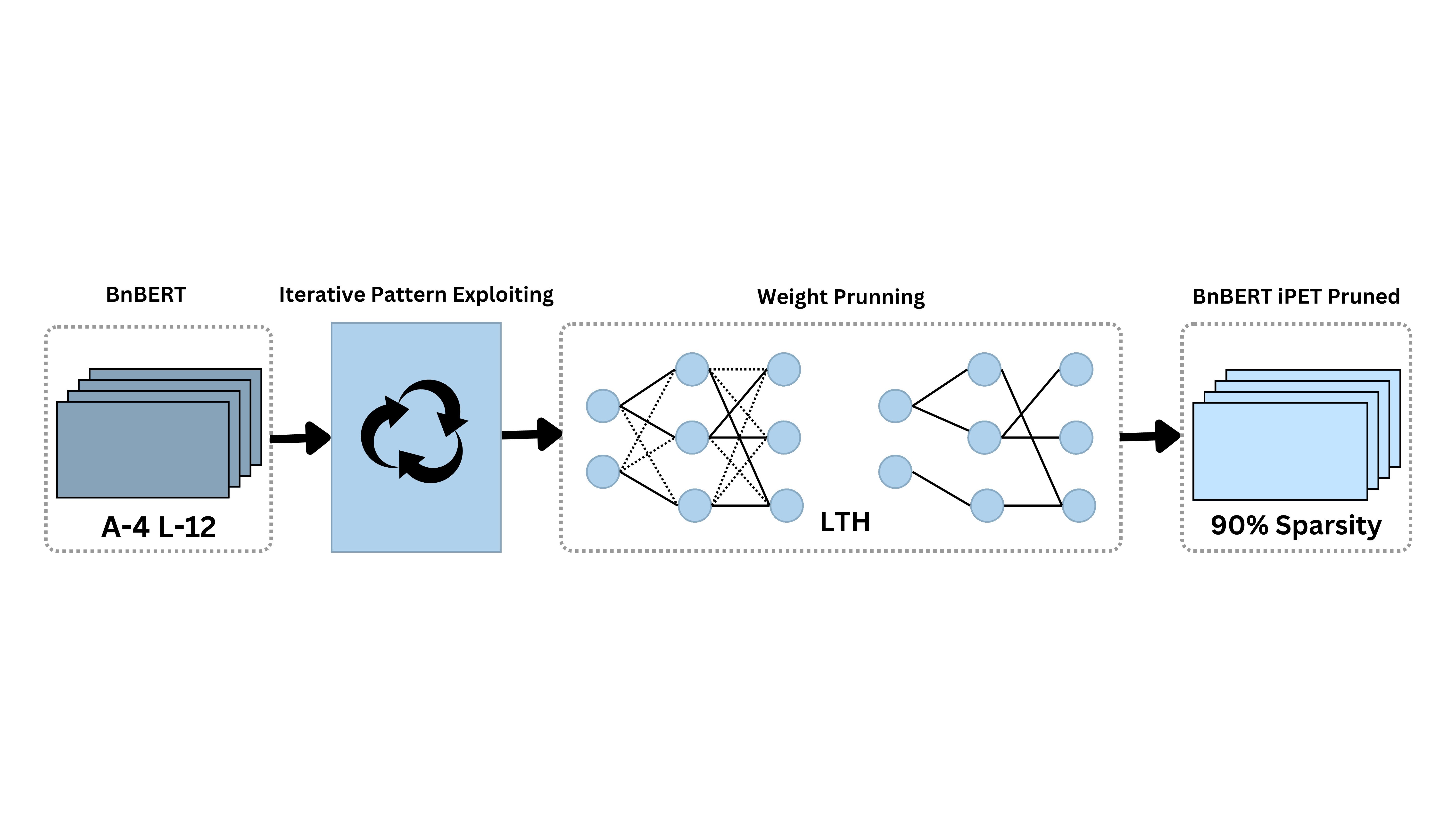}
\caption{Illustration of iterative pattern exploiting training and pruning of a tiny BERT model.}
\label{fig:methodology}
\end{figure}
\subsection{Iterative Pattern Exploiting Training}
We performed pattern exploiting training on a Bengali dataset using a pre-trained model with a customized architecture, following the technique of Schick and Sch\"utze \cite{b22}, as illustrated in Figure~\ref{fig:methodology}. In this technique, let $M$ be a masked language model (MLM) and $T$ its vocabulary, with $T^*$ denoting the set of all token sequences. For some $z \in T^*$ containing at least $k$ masks and $t \in T$, we denote by $q_M^k(t \mid z)$ the probability that $M$ assigns to $t$ at the $k$-th masked position in $z$; the model's logits before applying softmax are denoted by $s_M^k(t \mid z)$. The task is to map inputs $x \in X$ to outputs $y \in Y$, and for this PET needs a set of pattern-verbalizer pairs (PVPs). Each PVP $p = (P, v)$ consists of a pattern $P : X \to T^*$ that maps inputs to cloze questions containing a single mask, and a verbalizer $v : Y \to T$ that maps each output to a single token representing its task-specific meaning in the pattern. The core idea of PET is to derive the probability of $y$ being the correct output for $x$ from the probability of $v(y)$ being the ``correct'' token at the masked position in $P(x)$. Based on this intuition, a conditional probability distribution $q_p$ of $y$ given $x$ is defined as
\begin{equation}
q_p (y \mid x) = \frac{\exp s_p (y \mid x)}{\sum_{y'}\exp s_p (y' \mid x)}
\end{equation}
where $s_p (y \mid x) = s_M^1(v(y) \mid P( x ))$ is the raw score of $v(y)$ at the masked position in $P(x)$.
For a given task, identifying PVPs that perform well is challenging in the absence of a large development set. Therefore, PET enables a combination of multiple PVPs $\mathcal{P} = \{p_1, \ldots , p_n\}$ as follows: 
for each PVP $p$, an MLM is fine-tuned on training examples $(x, y)$ by minimizing the cross entropy between $y$ and $q_p (y \mid x)$, and the ensemble of fine-tuned MLMs is used to annotate a set of unlabeled examples; each unlabeled example $x \in X$ is annotated with soft labels based on the probability distribution below, where $w_p$ is a weighting term proportional to the accuracy achieved with $p$ on the training set before training. The resulting soft-labeled dataset is used to train a regular sequence classifier by minimizing the cross entropy between its output and $q_p$:
\begin{equation}
q_p (y \mid x) \propto \exp \sum_{p\in \mathcal{P}} w_p \cdot s_p (y \mid x)
\end{equation}
PET is an iterative technique in which successive generations of models are trained on increasingly large datasets that have been labeled by previous generations. This is accomplished in the following way: First, as in normal PET, an ensemble of MLMs is trained. For each model $M_i$, a random subset of other models is used to generate a new training set $t_i$ by assigning labels to those unlabeled examples for which the selected subset of models is most confident in its prediction. Each $M_i$  is then retrained on $t_i$; this process is repeated several times, each time increasing the number of examples in $t_i$  by a constant factor.

\subsection{Pruning}
For pruning, we used the Lottery Ticket Hypothesis technique. Given a network $f(x; \theta, \cdot)$, a subnetwork is a network $f(x; m \odot \theta, \cdot)$ with a pruning mask $m\in \{0,1\}^{d_1}$ (where $\odot$ is the element-wise product); that is, it is a copy of $f(x; \theta, \cdot)$ with some weights fixed to 0. Let $A_t^T  (f(x;\theta_i, \gamma_i)) $ be a training algorithm for a task $T$ that trains a network $ f(x;\theta_i, \gamma_i) $ on task $T$ for $t$ steps, creating network $ f(x;\theta_{i+t}, \gamma_{i+t}) $. Let $\theta_0$ be the BERT pre-trained weights, and let $\epsilon^T (f(x; \theta))$ be the evaluation metric of model $f$ on task $T$.\\
\textit{Matching subnetwork:} a subnetwork $f(x; m \odot \theta, \gamma)$ is matching for an algorithm $A_t^T$ if training $f(x; m \odot \theta, \gamma)$ with algorithm $A_t^T$ results in an evaluation metric on task $T$ no lower than that obtained by training $f(x; \theta_0,\gamma)$ with algorithm $A_t^T$:
\begin{equation}
\epsilon^T (A_t^T(f(x; m \odot \theta,\gamma)))\geq \epsilon^T (A_t^T (f(x;\theta_0,\gamma)))
\end{equation}
\textit{Winning ticket:} a subnetwork $f(x; m \odot \theta, \gamma)$ is a winning ticket for an algorithm $A_t^T$ if it is a matching subnetwork for $A_t^T$ and $\theta = \theta_0$.\\
\textit{Universal subnetwork:} a subnetwork $f(x; m \odot \theta, \gamma_{t_0} )$ is universal for tasks
$\{T_i\}_{i=1}^N$  if it is matching for each $A_{t_i}^{T_i}$ under appropriate, task-specific configurations of $\gamma_{T_i}$.

\subsection{Strategies}
\begin{figure}[!t]
\centering
\includegraphics[width=\columnwidth]{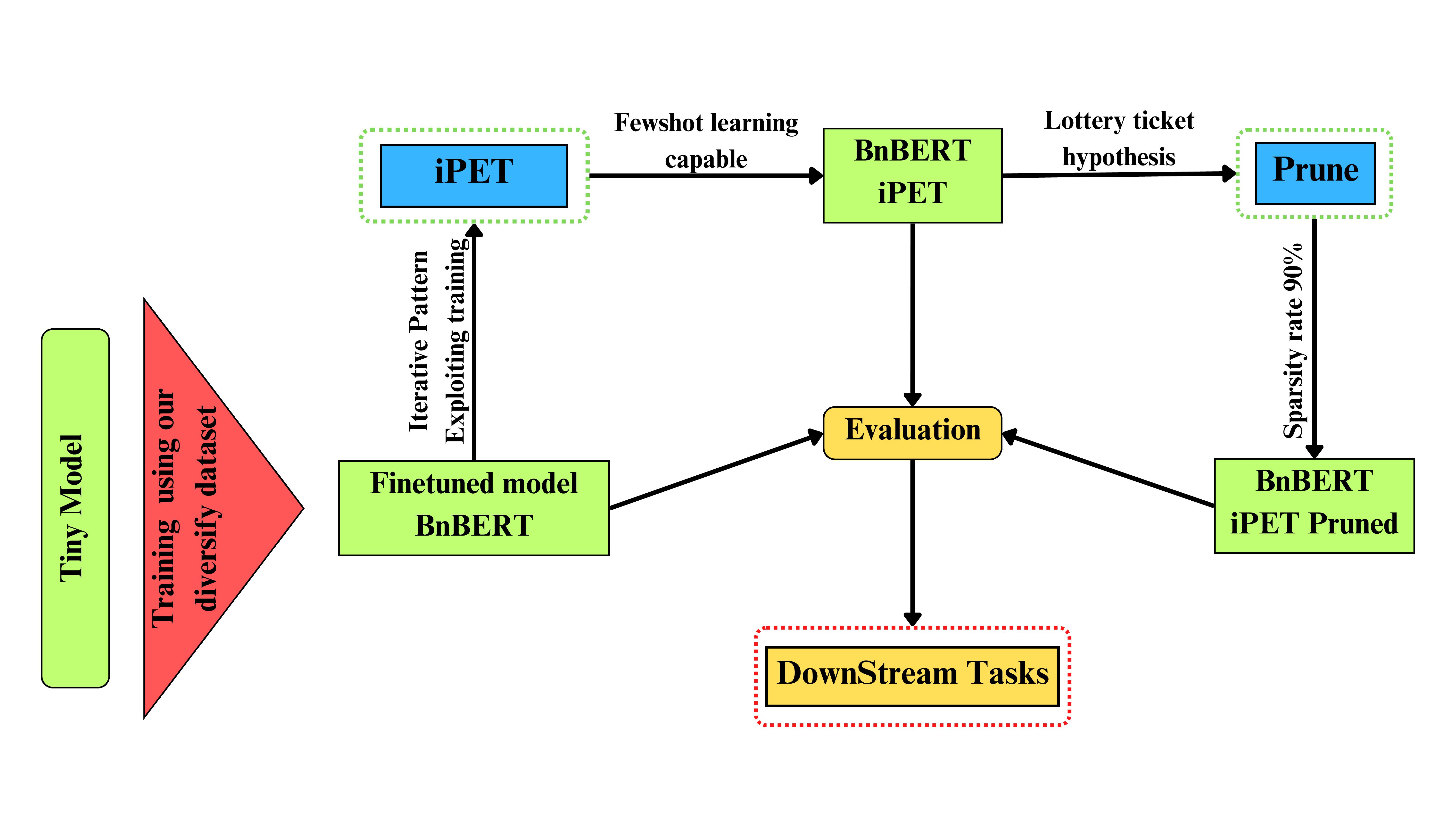}
\caption{An illustration of the work process. The performance of each model is measured on downstream tasks.}
\label{fig:workflow}
\end{figure}

Figure~\ref{fig:workflow} illustrates the steps of the model training and evaluation process used to assess the performance of our models on a variety of downstream tasks. In this experiment, we present three novel transformer-based models trained on a customized, diverse Bangla-language dataset. Our base model, BnBERT, is fine-tuned through unsupervised learning on a preprocessed diverse text dataset starting from the tiny Chinese ALBERT model \cite{b51}. To make the base model capable of few-shot learning, we applied iPET \cite{b22} to it. During the LM training of the BnBERT iPET model, we used a custom data processor and custom patterns for the iPET task, feeding the relevant dataset and declaring the custom pattern IDs. To make the BnBERT iPET model lighter while retaining performance, we dropped the unnecessary weights: we trained BnBERT iPET Pruned on the MLM objective using the Lottery Ticket Hypothesis \cite{b21} on the diverse dataset, customizing the training parameters to explore all iterations. During training, after every iteration, 10 percent of the network is dropped and the remaining weights are rewound for the next step, repeated until a 90 percent sparsity rate is reached. We experimented with a variety of downstream tasks to evaluate the effectiveness of the three models, BnBERT, BnBERT iPET, and BnBERT iPET Pruned, including measuring the perplexity of each model and of the dataset.

\section{Experiment}
\vspace{2mm}
\subsection{Dataset}
\begin{figure}[!t]
\centering
\includegraphics[width=\columnwidth]{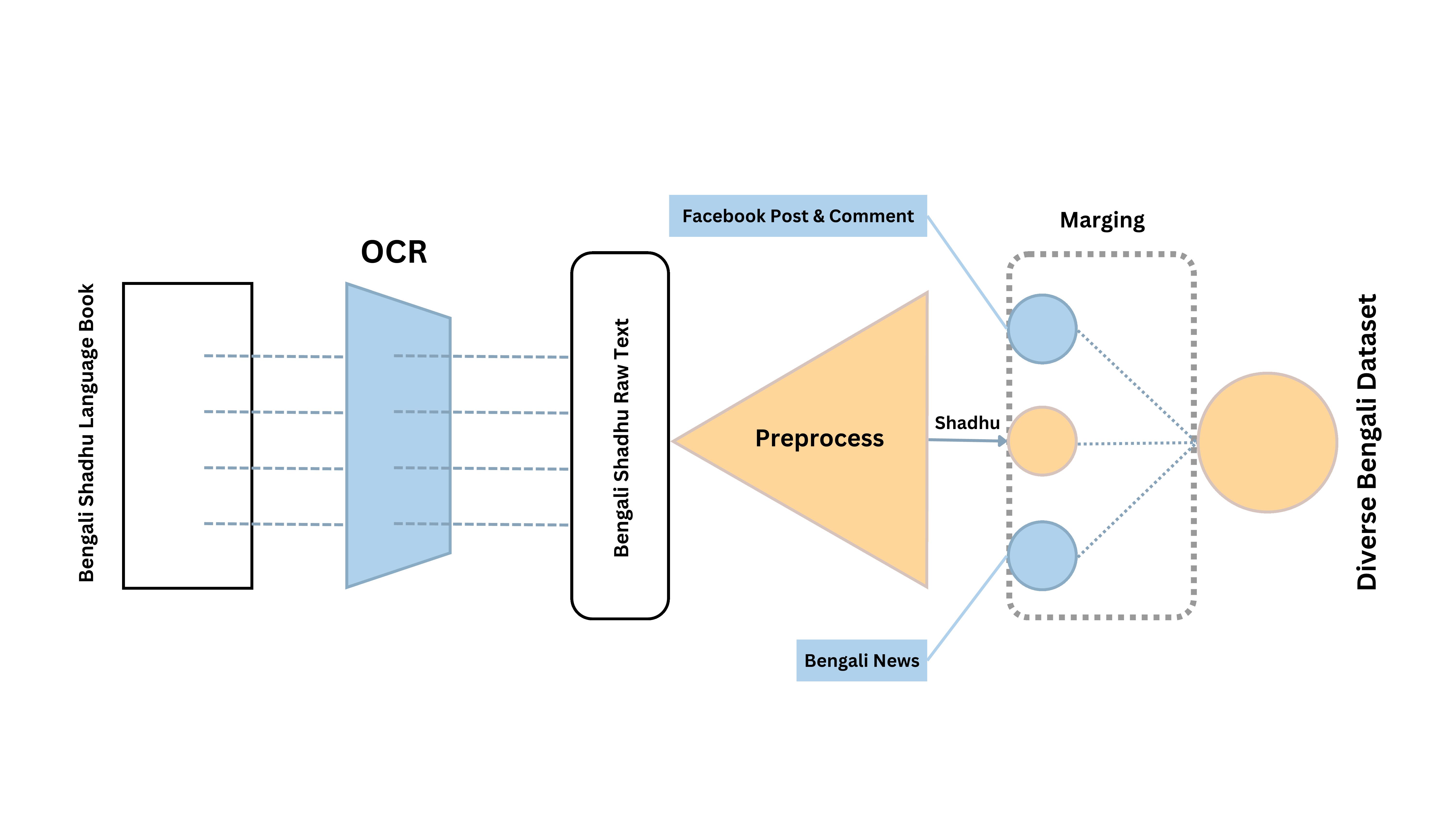}
\caption{The detailed process of data collection and construction of a diverse Bengali dataset.}
\label{fig:dataset}
\end{figure}
Pretraining a language model requires a large amount of high-quality text data. BERT, for example, is pretrained on 3.3 billion tokens from the English Wikipedia and the Books corpus. Bangla, however, is a resource-constrained language. In spite of this, we diversified our data across different available sources to make the pretrained model impactful and to cover all aspects of the language. We fine-tuned our pre-trained model using our own corpus compiled from several literature sources. To obtain our data, we collected a number of PDF files and extracted the Bangla text using OCR. We replaced contaminated and useless text with whitespace to eliminate polluted and unnecessary content from our data. The data was then preprocessed into BERT format, which is one sentence per line. After preprocessing, the overall size of our corpus is 80~MB, and it contains three types of data. To diversify the data, social media posts, Bangla newspapers, and Bangla Shadhu text were mixed. We divided the data into training and testing sets at an 8:2 ratio and fed them into our model. Figure~\ref{fig:dataset} shows the procedure for constructing the diverse dataset. 

\subsection{Downstream Task Datasets}
To evaluate the performance of our models, a number of downstream tasks are applied, using datasets from the study in \cite{b36}. Sentiment, emotion, authorship attribution, and news categorization are treated as text classification problems, where the text might be a document, an article, or a social media post, while POS tagging and punctuation restoration are examples of token classification problems. The models are fine-tuned end-to-end for these downstream tasks, utilizing the dataset throughout the entire network. 

\subsubsection{Emotion Classification}
\begin{figure}[!t]
\centering
\includegraphics[width=0.8\columnwidth]{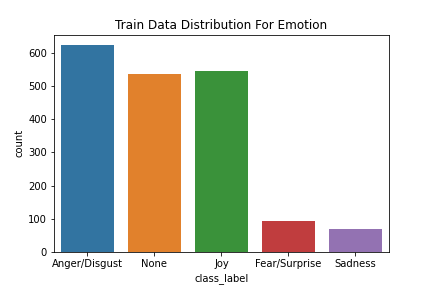}\\
\includegraphics[width=0.8\columnwidth]{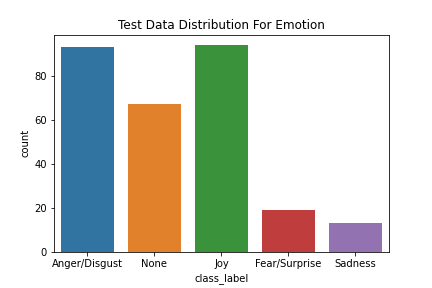}\\
\includegraphics[width=0.8\columnwidth]{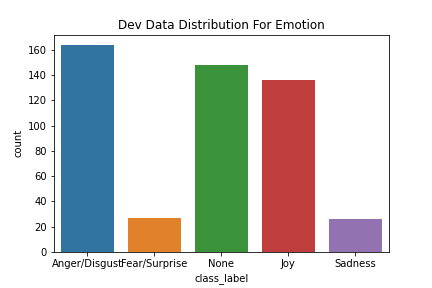}
\caption{Train, test, and development data distributions for the emotion dataset.}
\label{fig:dist-emotion}
\end{figure}

The emotion dataset includes five emotion labels: fear/surprise, joy, sadness, anger/disgust, and none\cite{b37, b38}. Following the data split method, 10\% of the dataset is reserved for testing, and the remainder is further divided into 20\% for development and 80\% for training. The dataset includes 2890 YouTube comments in Bangla, English, and romanized Bangla. The corresponding data distributions are shown in Figure~\ref{fig:dist-emotion}.\\

\subsubsection{Authorship Classification}
\begin{figure}[!t]
\centering
\includegraphics[width=0.8\columnwidth]{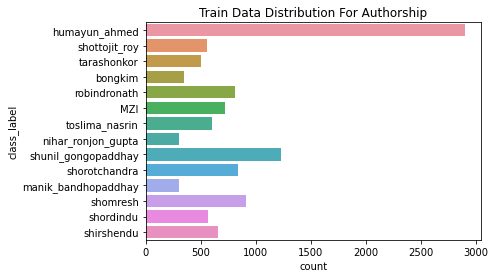}\\
\includegraphics[width=0.8\columnwidth]{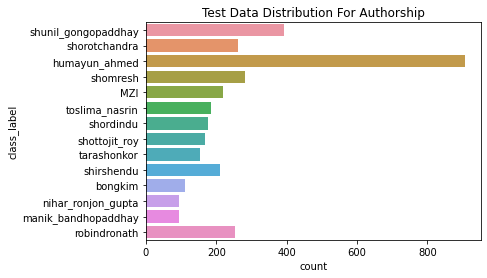}\\
\includegraphics[width=0.8\columnwidth]{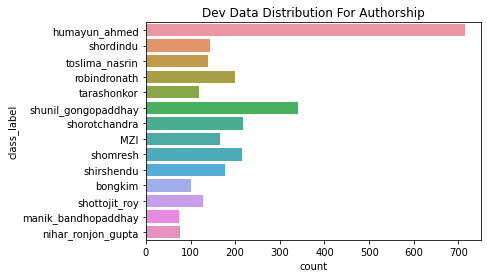}
\caption{Train, test, and development data distributions for the authorship dataset.}
\label{fig:dist-authorship}
\end{figure}

The authorship dataset includes the writings of 14 distinct writers\cite{b38,b39}. Each document in the dataset has a fixed length of 750 words. The data splits consist of 14047, 3511, and 750 writings for the train, development, and test divisions, respectively (Figure~\ref{fig:dist-authorship}). \\

\subsubsection{News Categorization}
\begin{figure}[!t]
\centering
\includegraphics[width=0.8\columnwidth]{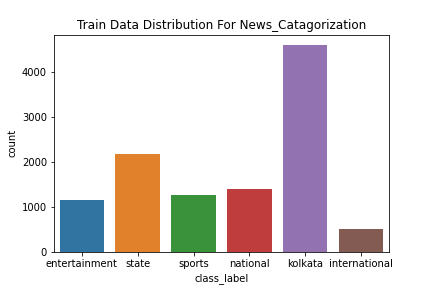}\\
\includegraphics[width=0.8\columnwidth]{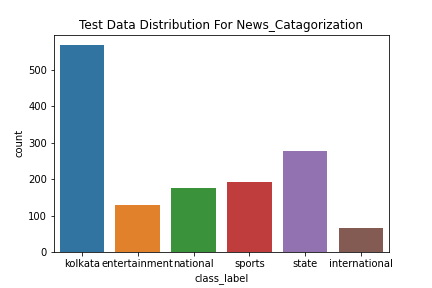}\\
\includegraphics[width=0.8\columnwidth]{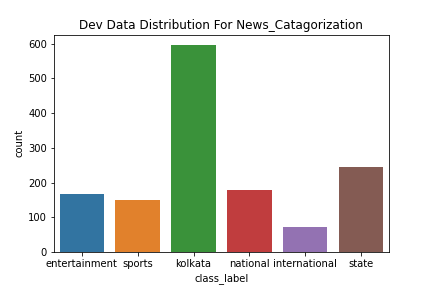}
\caption{Train, test, and development data distributions for the news categorization dataset.}
\label{fig:dist-newscatagorization}
\end{figure}

This news classification dataset, which has training, development, and test splits with 11109, 1408, and 1407 news articles, respectively, comprises six different class labels\cite{b38,b40}. The international news category has a low share of the distribution (Figure~\ref{fig:dist-newscatagorization}). \\

\subsubsection{Sentiment Classification}
For this research work, we combined and used separate publicly accessible sentiment analysis datasets \cite{b41}. Below, we provide a succinct overview of each dataset.\\

\begin{itemize}
    \item YouTube Comments Dataset\cite{b37}
\end{itemize}
\vspace{2mm}
The YouTube dataset has been created by combining a variety of comments posted on the site. There are 1660 comments for training, 273 comments for testing, and 297 comments for development in its three classes.\\

\begin{figure}[!t]
\centering
\includegraphics[width=0.8\columnwidth]{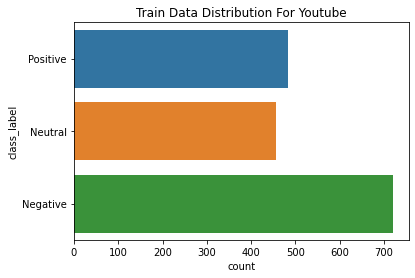}\\
\includegraphics[width=0.8\columnwidth]{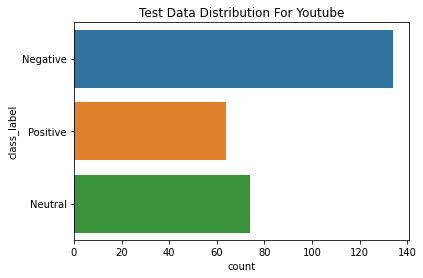}\\
\includegraphics[width=0.8\columnwidth]{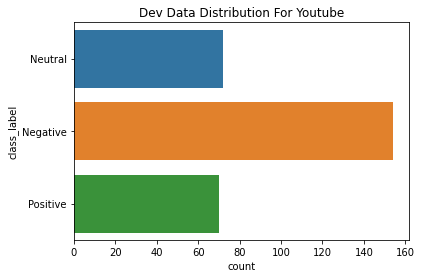}
\caption{Train, test, and development data distributions for the YouTube comments dataset.}
\label{fig:dist-youtube}
\end{figure}

\begin{itemize}
    \item BengFastText Dataset\cite{b42}
\end{itemize}
\vspace{2mm}
A combination of newspapers, TV news, books, blogs, and social media sites have been used to assemble the BengFastText dataset. The total quantity of data in this dataset is 7776, which is divided into 5253 training data, 1228 development data, and 1295 test data.\\

\begin{figure}[!t]
\centering
\includegraphics[width=0.8\columnwidth]{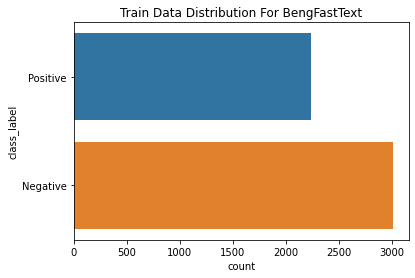}\\
\includegraphics[width=0.8\columnwidth]{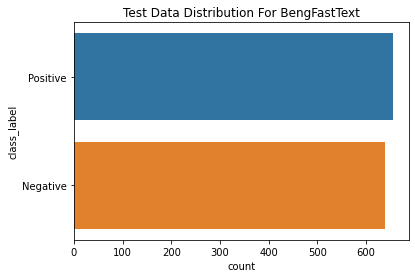}\\
\includegraphics[width=0.8\columnwidth]{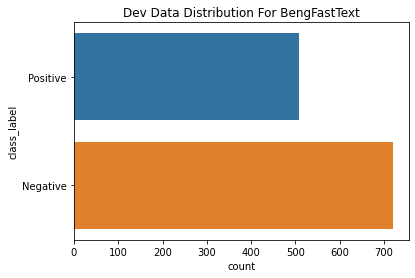}
\caption{Train, test, and development data distributions for the BengFastText dataset.}
\label{fig:dist-bengfasttext}
\end{figure}
\begin{itemize}
    \item Sail Dataset\cite{b43}
\end{itemize}
For the Shared Task on Sentiment Analysis in Indian Languages (SAIL), a dataset named SAIL was constructed, most of which are tweets. The dataset includes a total of 1000 tweets, which are broken down into 697 posts for training, 204 posts for testing, and 99 posts for development, respectively.\\

\begin{figure}[!t]
\centering
\includegraphics[width=0.8\columnwidth]{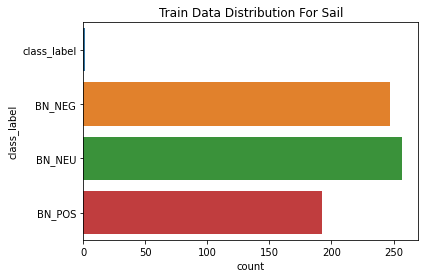}\\
\includegraphics[width=0.8\columnwidth]{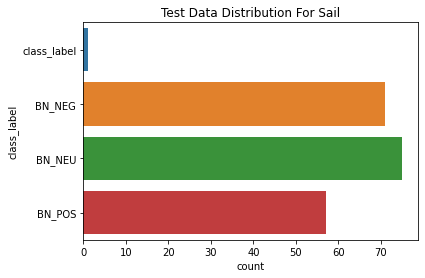}\\
\includegraphics[width=0.8\columnwidth]{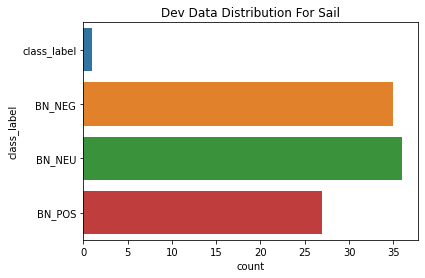}
\caption{Train, test, and development data distributions for the SAIL dataset.}
\label{fig:dist-sail}
\end{figure}
\begin{itemize}
    \item ABSA Cricket Dataset\cite{b44}
\end{itemize}
\vspace{-2mm}
This dataset, which was created using data from Facebook, BBC Bangla, and Prothom Alo, comprises a total of 2688 entries, of which 1943 are for training, 373 are for development, and 372 are for testing.

\begin{figure}[!t]
\centering
\includegraphics[width=0.8\columnwidth]{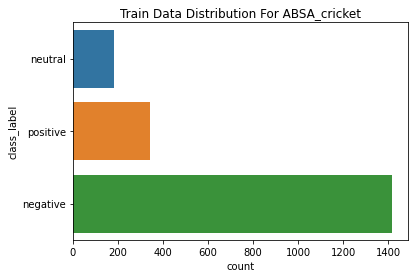}\\
\includegraphics[width=0.8\columnwidth]{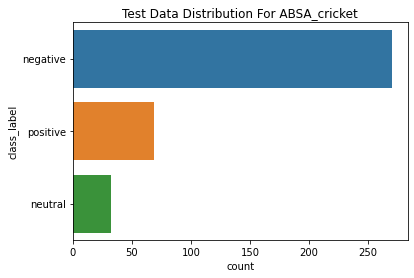}\\
\includegraphics[width=0.8\columnwidth]{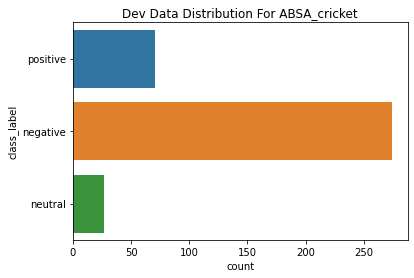}
\caption{Train, test, and development data distributions for the ABSA cricket dataset.}
\label{fig:dist-absacricket}
\end{figure}

\begin{itemize}
    \item ABSA Restaurant\cite{b44}    
\end{itemize}
\vspace{1mm}
To perform aspect-based sentiment analysis in Bangla, the ABSA dataset was assembled. Only the restaurant category of data is contained in the dataset. The dataset includes 225 data for development, 209 data for testing, and 1188 data for training. This dataset contains three classes.

\begin{figure}[!t]
\centering
\includegraphics[width=0.8\columnwidth]{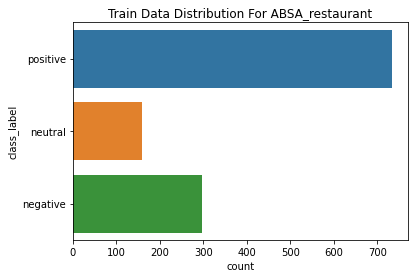}\\
\includegraphics[width=0.8\columnwidth]{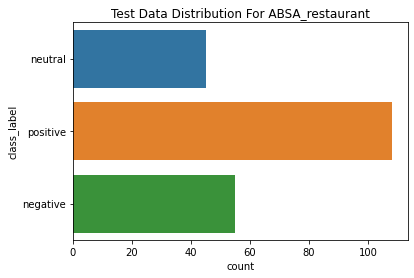}\\
\includegraphics[width=0.8\columnwidth]{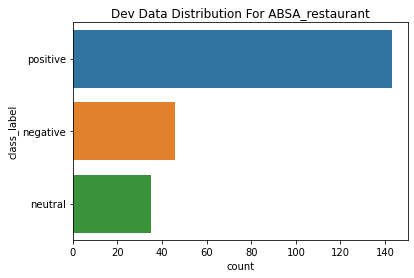}
\caption{Train, test, and development data distributions for the ABSA restaurant dataset.}
\label{fig:dist-absarestaurant}
\end{figure}
\begin{itemize}
    \item CogniSenti Dataset\cite{b41}
\end{itemize}
\vspace{2mm}
The CogniSenti dataset, which is composed of social media posts, has 5,628 tweets from Twitter and 942 posts from Facebook. The data is divided into training, development, and test sets of 4,599, 985, and 986 samples, respectively.
\vspace{2mm}
\begin{figure}[!t]
\centering
\includegraphics[width=0.8\columnwidth]{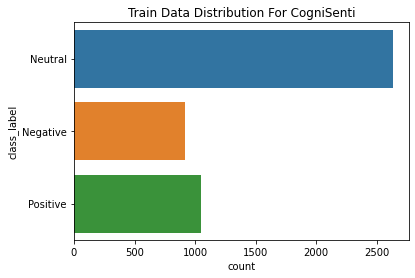}\\
\includegraphics[width=0.8\columnwidth]{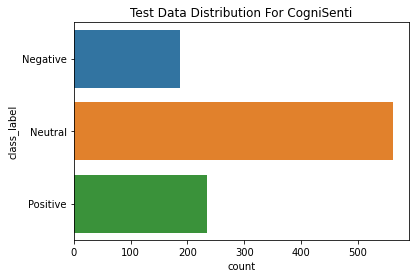}\\
\includegraphics[width=0.8\columnwidth]{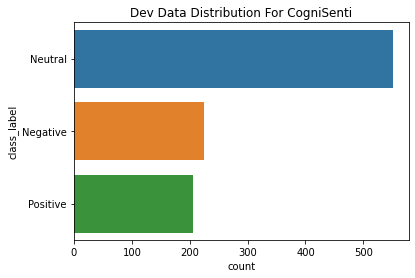}
\caption{Train, test, and development data distributions for the CogniSenti dataset.}
\label{fig:dist-cognisenti}
\end{figure}
\begin{itemize}
    \item Combined      
\end{itemize}
This dataset separately combines all training datasets, test datasets, and development datasets of sentiment classification. It includes 4807 training data, 1031 test data, and 1031 development data.

\begin{figure}[!t]
\centering
\includegraphics[width=0.8\columnwidth]{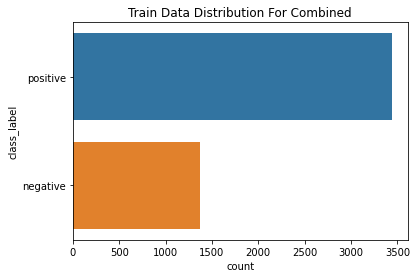}\\
\includegraphics[width=0.8\columnwidth]{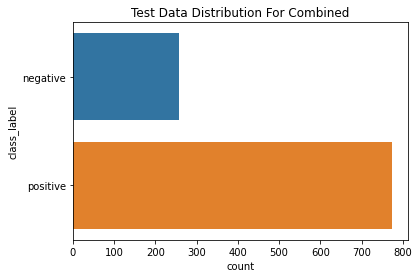}\\
\includegraphics[width=0.8\columnwidth]{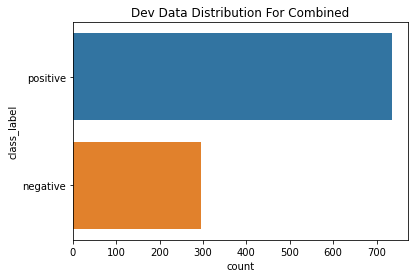}
\caption{Train, test, and development data distributions for the combined sentiment dataset.}
\label{fig:dist-combined}
\end{figure}

\subsubsection{POS Tagging}
For training and evaluating the models on the POS tagging task, we used three datasets: the LDC Corpus\cite{b45,b46}, the IITKGP POS Tagged Corpus\cite{b47}, and the CRBLP POS Tagged Corpus\cite{b48}. We used a three-step procedure for training and testing the models.\\
\begin{itemize}
   \item LDC Corpus
\end{itemize}
In the first step, we used the LDC Corpus, which has a total of 7,393 sentences, or 102,937 tokens. It was developed by Microsoft Research (MSR), India, for linguistic research. The corpus text was compiled from blogs, Wikipedia articles, and other sources to provide diversity.\\
\begin{itemize}
\item LDC+IITKGP Corpus
\end{itemize}
The IITKGP POS Tagged Corpus and LDC Corpus were integrated in the next phase. There are 72,400 tokens and 5,473 phrases in the IITKGP POS Tagged Corpus. Microsoft Research created this corpus in conjunction with IIT Kharagpur and numerous other Indian institutes.\\
\begin{itemize}
\item LDC+IITKGP+CRBLP Corpus
\end{itemize}
The last step combined the LDC+IITKGP corpus above with the CRBLP POS Tagged Corpus, in which 20,000 tokens are distributed across 1,176 sentences.\\

\subsubsection{Punctuation Restoration}
The train, development, and test splits for the punctuation restoration task were created using a publicly available corpus of Bangla articles\cite{b49, b50}. The authors created two additional test datasets from ASR output and manual transcripts, compiled from 65 minutes of voice snippets taken from four short tales in Bangla. There are four labels: three punctuation marks (comma, period, and question mark) plus the ``other'' token class. This dataset consists of 1,379,986 tokens for train, 179,371 tokens for dev, and 87,721, 6,821, and 6,417 tokens for Test (news), Test (Ref.), and Test (ASR), respectively. \\

\subsection{Configurations}
The fine-tuned model, which has a vocabulary of 30522 tokens, a hidden size of 312, and 4 hidden layers, is considered our base model. The two other models share the same configuration but utilize the iPET and pruning methodologies. We selected seven models from the literature to compare against these three models. Table~\ref{tab:config} shows the configurations of the chosen models.\\

\subsection{Metrics}
Perplexity is a metric that measures how effectively a probability model predicts a given sample, and it is one common technique for evaluating language models in natural language processing. A good language model is one that predicts previously unseen data well. We evaluate our models based on perplexity:
\begin{equation}
PP(W) = \sqrt[N]{\prod_{i=1}^N\frac{1}{P(W_i|W_{i-1})}}
\end{equation}
We measured the perplexity of our dataset using this method, both overall and for each category of the dataset. The results are shown in Table~\ref{tab:perp-data}.\\
\begin{table}[h!]
\centering
\caption{Perplexity of the diverse dataset}
\label{tab:perp-data}
\begin{tabular}{*{3}{r}}
 \hline
 Dataset & Eval loss & Perplexity \\ 
 \hline
 Full Dataset & 2.28 & 9.43 \\ [1ex]
 Sadhu & 2.29 & 9.88 \\ [1ex]
 Social Media & 2.36 & 10.52\\ [1ex]
 Newspaper & 2.24 & 9.29 \\ [1ex] 
 \hline
\end{tabular}

\end{table}


\begin{table}[h!]
\caption{Perplexity of each model based on training loss}
\label{tab:perp-model}
\centering
\begin{tabular}{*{2}{r}}
 \hline
 Model & Perplexity \\ [1ex]
 \hline
 BnBERT(ours) & 3.8722 \\ [1ex]
 BnBERT iPET(ours) & 1.0460 \\ [1ex]
 BnBERT iPET Pruned(ours) & 46.8558 \\ [1ex]
 \hline
\end{tabular}
\end{table}


\begin{table*}
\centering
\caption{Configurations of our models and of the models compared against them}
\label{tab:config}
\begin{tabular}{*{5}{r}}
  \toprule
  Model  & Attention heads & Hidden layers & Hidden size & Vocab size \\
  \midrule
  BnBERT(ours) & 12 & 4 & 312 & 30522 \\
  BnBERT iPET(ours) & 12 & 4 & 312 & 30522 \\
  BnBERT iPET Pruned(ours) & 12 & 4 & 312 & 30522 \\
  BanglaBERT\cite{b27} & 12 & 12 & 768 & 32000 \\
  Bangla-Electra\cite{b12} & 4 & 12 & 256 & 29898 \\
  Indic-BERT\cite{b12} & 12 & 12 & 768 & 100000 \\
  BERT-bn\cite{b12} & 12 & 12 & 768 & 102025 \\
  XLM-RoBERTa\cite{b12} & 16 & 24 & 1024 & 250002 \\
  Indic-DistilBERT\cite{b12} & 12 & 6 & 768 & 30522 \\
  DistilBERT-m\cite{b12} & 12 & 6 & 768 & 119547 \\
  \bottomrule                             
\end{tabular}
\end{table*}


\begin{table*}
\centering
\caption{Emotion Classification}
\label{tab:emotion}
\begin{tabular}{*{5}{r}}
  \toprule
  Model & Acc & P & R & F1 \\
  \midrule
  BnBERT(ours) & 45.51 & 43.06 & 45.51 &\textbf{41.21} \\
  BnBERT iPET(ours) & 43.71 & 41.08 & 43.71 & 39.84\\
  BnBERT iPET Pruned(ours) & 50.30 & 46.86 & 50.30 & \textbf{48.10}\\
  Bangla Electra\cite{b12} & 43.8 & 38.3 & 43.8 & 36.3\\
  Indic-BERT\cite{b12} &50.6 &52.1 &50.6 &49.1\\
  BERT-bn\cite{b12} &49.1 &46.7 &49.1 &46.9\\
  BanglaBERT\cite{b27} &71.05& 66.46 &71.05 &68.62\\

  \bottomrule                             
\end{tabular}
\end{table*}


\begin{table*}
\centering
\caption{Authorship Classification}
\label{tab:authorship}
\begin{tabular}{*{5}{r}}
  \toprule
  Model & Acc & P & R & F1 \\
  \midrule
  BnBERT(ours) &97.11 &97.14 &97.11 & \textbf{97.11}\\
BnBERT iPET(ours) &97.72 &97.76 &97.72 & \textbf{97.73}\\
BnBERT iPET Pruned(ours) &97.67 &97.69 &97.67 & \textbf{97.68}\\
XLM-RoBERTa\cite{b12} &93.8 &94.1 &93.8 &93.8\\
Indic-BERT\cite{b12} &95.2 &95.3 &95.2 &95.2\\
BERT-bn\cite{b12} &90.2 &90.3 &90.2 &90.2\\
BanglaBERT\cite{b27} &98.85 &97.71 &97.88 &98.71\\

  \bottomrule                             
\end{tabular}
\end{table*}


\begin{table*}
\centering
\caption{News Categorization}
\label{tab:news}
\begin{tabular}{*{5}{r}}
  \toprule
  Model & Acc & P & R & F1 \\
  \midrule
  BnBERT(ours) &82.51 &83.03 &82.51 &81.50\\
BnBERT iPET(ours) &84.29 &84.35 &84.29 & \textbf{84.08}\\
BnBERT iPET Pruned(ours) &88.48 &88.24 &88.48 & \textbf{88.19}\\
Bangla Electra\cite{b12} &80.4 &78.5 &80.4 &79.2\\
Indic-DistilBERT\cite{b12} &89.0 &90.2 &89.0 &89.4\\
DistilBERT-m\cite{b12} &79.5 &79.4 &79.5 &79.0\\
BanglaBERT\cite{b27} &94.52 &94.55 &94.52 &94.53\\

  \bottomrule                             
\end{tabular}
\end{table*}


\begin{table*}
\begin{center}
\caption{Punctuation Restoration}
\label{tab:punct}
\begin{tabular}{*{10}{c} }
\toprule
    & \multicolumn{3}{c}{News}
            & \multicolumn{3}{c}{Ref.}
                            & \multicolumn{3}{c}{ASR}                \\
                            \midrule
    &   P  &   R  &   F1  &   P  &   R  &   F1  &   P  &   R  &   F1   \\
    \hline\\
BnBERT
&94.32
&98.38
&\textbf{96.31}
&85.04
&97.11
&\textbf{90.67}
&86.55
&96.71
&\textbf{91.35}\\
BnBERT iPET
&87.94
&95.20
&93.76
&79.0
&90.13
&88.27
&80.46
&90.11
&89.17\\
BnBERT iPET Pruned
&94.55
&98.23
&\textbf{96.36}
&86.99
&96.19
&\textbf{91.36}
&88.51
&95.73
&\textbf{91.98}\\
XLM-RoBERTa\cite{b12}
&87.8
&86.2
&87.0
&67.6
&70.2
&68.6
&60.3
&66.4
&63.2\\
Indic-BERT\cite{b12}
&73.9
&70.5
&72.2
&60.7
&54.1
&57.2
&55.9
&53.4
&54.7\\
Bangla Electra\cite{b12}
&64.8
&49.7
&56.3
&59.5
&39.2
&47.2
&54.8
&38.7
&45.3\\
BanglaBERT\cite{b27}
&94.77
&98.21
&96.46
&68.07
&40.64
&50.86
&68.93
&56.21
&61.93\\

\bottomrule 
\end{tabular}
    \end{center}
\end{table*}

\begin{table*}
\begin{center}
\caption{POS Tagging}
\label{tab:pos}
\begin{tabular}{*{13}{c} }
\toprule
    & \multicolumn{4}{c}{BnBERT(ours)}
            & \multicolumn{4}{c}{BnBERT iPET(ours)}
                            & \multicolumn{4}{c}{BnBERT iPET Pruned(ours)}                \\
                            \midrule
    &   Acc  &   P  &   R  &   F1  &   Acc  &   P  &   R  &   F1  &   Acc  &   P  &   R  &   F1   \\
    \hline\\
LDC
&75.30
&69.04
&65.06
&65.78
&75.87
&69.08
&66.80
&67.37
&81.40
&74.94
&73.39
&73.85\\
LDC+IITKGP
&79.04
&72.81
&70.66
&71.09
&78.87
&72.59
&70.82
&71.28
&83.27
&77.47
&76.72
&\textbf{76.93}\\
LDC+IITKGP+CRBLP
&78.08
&70.91
&69.38
&69.30
&77.73
&70.36
&69.15
&69.24
&82.01
&75.51
&74.78
&\textbf{82.01}\\

\hline\\
    & \multicolumn{4}{c}{Bangla Electra\cite{b12}}
            & \multicolumn{4}{c}{DistilBERT-m\cite{b12}}
                            & \multicolumn{4}{c}{BERT-bn\cite{b12}}             \\
                            \midrule
    &   Acc  &   P  &   R  &   F1  &   Acc  &   P  &   R  &   F1  &   Acc  &   P  &   R  &   F1   \\
    \hline\\
LDC
&79.0
&73.4 &71.1
&72.2 &83.9 
&78.8 &78.0 &78.4
&85.8 &81.5
&80.7 &81.1\\
LDC+IITKGP
&80.2 &74.8
&72.2 &73.7
&83.7 &78.5
&77.9 &78.2
&85.5 &81.5
&80.9 &81.2\\
LDC+IITKGP+CRBLP
&80.4 &75.1
&73.0 &74.7
&83.8 &78.4
&78.0 &78.20
&85.4 &81.0
&80.2 &80.6\\

\hline\\
   & \multicolumn{4}{c}{BanglaBERT\cite{b27}}
            & \multicolumn{4}{c}{}
                            & \multicolumn{4}{c}{}                \\
                            \midrule
    &   Acc  &   P  &   R  &   F1  \\
    \hline\\
LDC
&91.54
&88.40
&89.04
&88.67\\
LDC+IITKGP
&92.45
&89.75
&90.24
&89.93\\
LDC+IITKGP+CRBLP
&92.08
&88.97
&89.59
&89.22\\

\bottomrule 
\end{tabular}
    \end{center}
\end{table*}

\begin{table*}
\begin{center}
\caption{Sentiment Classification}
\label{tab:sentiment}
\begin{tabular}{*{13}{c} }
\toprule
   & \multicolumn{4}{c}{BnBERT(ours)}
            & \multicolumn{4}{c}{BnBERT iPET(ours)}
                            & \multicolumn{4}{c}{BnBERT iPET Pruned(ours)}                \\
                            \midrule
    &   Acc  &   P  &   R  &   F1  &   Acc  &   P  &   R  &   F1  &   Acc  &   P  &   R  &   F1   \\
    \hline\\
YouTube comments &63.85 &67.83 &63.85 &\textbf{65.28} &57.43 &70.29 &57.43 &58.76 &64.52 &65.42 &64.52 &64.91 \\
BengFastText &81.41 &81.81 &81.41 &\textbf{81.51} &79.86 &80.02 &79.86 &79.47 &79.62 &80.38 &79.62 &\textbf{79.76}\\
Sail &54.08 &66.25 &54.08 &\textbf{55.33} &46.93 &55.70 &46.93 &45.26 &54.08 &52.99 &54.08 &52.60 \\
ABSA Cricket &75.53 &69.48 &75.53 &\textbf{72.35} &77.41 &70.46 &77.41 &72.17 &72.31 &71.66 &72.31 &\textbf{71.62}\\
ABSA Restaurant &58.92 &48.49 &58.92 &53.09 &58.03 &61.35 &58.03 &57.09 &60.71 &58.05 &60.71 &\textbf{58.32}\\
CogniSenti &65.85 &58.23 &65.85 &58.15 &64.62 &49.45 &64.62 &56.00 &65.54 &66.02 &65.54 &\textbf{60.22}\\
Combined &77.91 &79.21 &77.91 &77.62 &79.52 &79.75 &79.52 &\textbf{79.10} &78.52 &78.49 &78.52 &\textbf{78.41}\\
\hline\\
    & \multicolumn{4}{c}{Bangla Electra\cite{b12}}
            & \multicolumn{4}{c}{DistilBERT-m\cite{b12}}
                            & \multicolumn{4}{c}{BERT-bn\cite{b12}}                \\
                            \midrule
    &   Acc  &   P  &   R  &   F1  &   Acc  &   P  &   R  &   F1  &   Acc  &   P  &   R  &   F1   \\
    \hline\\
YouTube comments &67.4 &66.2 &67.4 &66.6 &70.0 &70.4 &70.0 &70.1 &73.3 &72.6 &73.3 &72.9\\
BengFastText &68.7 &73.0 &68.7 &66.8 &69.1 &74.6 &69.1 &66.9 &69.0 &75.5 &69.0 &66.5\\
Sail &53.3 &51.9 &53.3 &49.8 &57.3 &57.0 &57.3 &57.0 &60.7 &59.8 &60.7 &59.5\\
ABSA Cricket &71.3 &63.5 &71.3 &67.0 &74.5 &66.5 &74.5 &69.9 &71.1 &67.8 &71.1 &69.2\\
ABSA Restaurant &50.2 &35.7 &50.2 &40.9 &60.3 &57.7 &60.3 &57.9 &60.7 &59.3 &60.7 &58.5\\
CogniSenti &63.9 &58.3 &63.9 &59.3 &59.2 &58.6 &59.2 &58.9 &62.4 &62.1 &62.4 &62.2\\
Combined &72.0 &72.4 &72.0 &71.8 &74.3 &74.8 &74.3 &74.2 &79.3 &79.3 &79.3 &79.2\\
\hline\\
    & \multicolumn{4}{c}{BanglaBERT\cite{b27}}
            & \multicolumn{4}{c}{}
                            & \multicolumn{4}{c}{}                \\
                            \midrule
    &   Acc  &   P  &   R  &   F1  \\
    \hline\\
YouTube comments &75.33 &75.27 &75.33 &75.21\\
BengFastText &84.75 &84.70 &84.75 &84.69\\
Sail &71.42 &72.11 &71.42 &70.80\\
ABSA Cricket &80.64 &82.50 &80.64 &76.02\\
ABSA Restaurant &72.76 &71.87 &72.76 &72.20\\
CogniSenti &70.03 &72.58 &70.03 &69.14\\
Combined &95.43 &95.76 &95.43 &95.50\\

\bottomrule 
\end{tabular}
    \end{center}
\end{table*}

\section{Results}
\vspace{2mm}
Following our proposed strategies, the perplexity of each of the three models was measured and is shown in Table~\ref{tab:perp-model}. Applying the iPET methodology to the base model yields better perplexity than both the base model and the pruned model. The same downstream tasks were applied to each model to evaluate how well it performs. All of the same downstream tasks were also carried out on BanglaBERT for comparison with our models. Additionally, we compared the performance of our models with other well-known models from the literature\cite{b36}. We used the weighted average for precision, recall, and F1 to evaluate the performance of our models, since it can handle the problem of dataset imbalance.

\subsection{Emotion Classification}
Table~\ref{tab:emotion} shows how well our models performed at this specific task. According to the F1 score, our BnBERT iPET Pruned and BnBERT models outperform Bangla Electra at emotion classification. Furthermore, our BnBERT iPET Pruned model performs better than BERT-bn and nearly as well as Indic-BERT.\\

\subsection{Authorship Classification}
Table~\ref{tab:authorship} compares the performance of our models against that of XLM-RoBERTa, Indic-BERT, and BERT-bn. BnBERT, BnBERT iPET, and BnBERT iPET Pruned achieve F1 scores of 97.11\%, 97.73\%, and 97.68\%, respectively. Additionally, our models provide outcomes equivalent to BanglaBERT.\\

\subsection{News Categorization}
Table~\ref{tab:news} contains the findings for the news categorization task. Our BnBERT iPET Pruned model works almost as well as Indic-DistilBERT for categorizing news, and all three of our models outperform Bangla Electra and DistilBERT-m. In terms of accuracy, BnBERT iPET Pruned is also able to compete with the larger BanglaBERT model.\\

\subsection{POS Tagging}
Table~\ref{tab:pos} demonstrates the performance of our models in comparison to Bangla Electra, DistilBERT-m, BERT-bn, and BanglaBERT. The accuracy and F1 score of BnBERT iPET Pruned are comparable with those of Bangla Electra and DistilBERT-m. It achieves an F1 score of 82.01\% on the three merged datasets, whereas Bangla Electra achieves 74.7\% and DistilBERT-m achieves 78.20\% on the same merged datasets.\\

\subsection{Sentiment Classification}
Table~\ref{tab:sentiment} summarizes the sentiment classification results of several transformer models for both the individual and combined datasets. BnBERT, BnBERT iPET, and BnBERT iPET Pruned perform better than Bangla Electra and DistilBERT-m in terms of accuracy and F1 score on the combined dataset: Bangla Electra and DistilBERT-m reach 72.0\% and 74.3\% accuracy, respectively, whereas BnBERT iPET reaches 79.52\%. BERT-bn and our models are also comparable on the aggregated dataset. On this specific task, BanglaBERT outperforms all the other models.\\

\subsection{Punctuation Restoration}
For this task, we consider only the overall results, ignoring the no-punctuation entries (O tokens), when presenting the punctuation restoration outcomes in Table~\ref{tab:punct}. BnBERT, BnBERT iPET, and BnBERT iPET Pruned outperform XLM-RoBERTa, Indic-BERT, Bangla Electra, and BanglaBERT across the three test sets on this task. On the News, Ref., and ASR test sets, BnBERT iPET Pruned achieves F1 scores of 96.36\%, 91.36\%, and 91.98\%, respectively. \\

\section{Inference Time Comparison}
\vspace{2mm}
According to the study in \cite{b16}, transformer models have large computational complexity, and their training and inference times are significantly higher than those of traditional algorithms (such as SVM and RF). The combined sentiment, authorship classification, emotion classification, and news categorization datasets were used to measure the training and inference times of our BnBERT iPET Pruned model. The training times for these tasks were 1 hour 40 minutes 39 seconds, 1 hour 8 minutes 34 seconds, 3 minutes 14 seconds, and 58 minutes 8 seconds, respectively, and inference took 15, 16, almost 4, and 5 seconds, respectively. SVM required 2 hours 32 minutes 30 seconds for training and 4 seconds for inference. For XLM-RoBERTa, the training phase took 3 hours 42 minutes 41 seconds, while inference took 1 minute 21 seconds. The inference time of our model was 3.75 times higher than that of SVM but 5.4 times lower than that of XLM-RoBERTa. Our research shows that, in terms of inference time, the BnBERT iPET Pruned model is able to compete with large models like XLM-RoBERTa and BanglaBERT on several downstream tasks. This model was fine-tuned from a small model with 12 attention heads and 4 layers using our diversified dataset.\\

\section{Conclusion}
As a consequence of our research, we found that a smaller model with a small number of layers can be a few-shot learner. This few-shot-learned model can outperform the pretrained base model in perplexity thanks to iterative pattern exploiting training. Although this training has minimal influence on downstream performance, it does allow the model to learn from a few shots. The Lottery Ticket Hypothesis progressively decreases the size of the deep neural network, removing superfluous weights until reaching 90\% sparsity and making the model lighter, greener, and more efficient to train and deploy. As a result, the model is comparable with complex state-of-the-art models on downstream tasks while having just 10\% of the weights of the main pretrained model. In future work, we plan to further increase the efficacy and efficiency of this strategy. 



\end{document}